\documentclass{ecai} 

\usepackage{latexsym}
\usepackage{amssymb}
\usepackage{amsmath}
\usepackage{amsthm}
\usepackage{booktabs}
\usepackage{enumitem}
\usepackage{graphicx}
\usepackage{color}
\usepackage{multirow}
\usepackage{algorithm}
\usepackage{algpseudocode}
\usepackage{amsmath}
\usepackage{url}

\newcommand{\BibTeX}{B\kern-.05em{\sc i\kern-.025em b}\kern-.08em\TeX}

\makeatletter
\def\blfootnote{\xdef\@thefnmark{}\@footnotetext}
\makeatother

\begin{document}


\begin{frontmatter}


\paperid{9460} 


\title{Uncertainty-Aware Knowledge Transformers for Peer-to-Peer Energy Trading with Multi-Agent Reinforcement Learning}


\author[A]{\fnms{Mian Ibad Ali}~\snm{Shah}\orcid{0009-0008-7288-9757}\thanks{Corresponding Author. Email: m.shah7@universityofgalway.ie}}
\author[A]{\fnms{Enda}~\snm{Barrett}}
\author[A]{\fnms{Karl}~\snm{Mason}}

\address[A]{School of Computer Science, University of Galway, Ireland}


\begin{abstract}
This paper presents a novel framework for Peer-to-Peer (P2P) energy trading that integrates uncertainty-aware prediction with multi-agent reinforcement learning (MARL), addressing a critical gap in current literature. In contrast to previous works relying on deterministic forecasts, the proposed approach employs a heteroscedastic probabilistic transformer-based prediction model called Knowledge Transformer with Uncertainty (KTU) to explicitly quantify prediction uncertainty, which is essential for robust decision-making in the stochastic environment of P2P energy trading. The KTU model leverages domain-specific features and is trained with a custom loss function that ensures reliable probabilistic forecasts and confidence intervals for each prediction. Integrating these uncertainty-aware forecasts into the MARL framework enables agents to optimize trading strategies with a clear understanding of risk and variability. Experimental results show that the uncertainty-aware Deep Q-Network (DQN) reduces energy purchase costs by up to 5.7\% without P2P trading and 3.2\% with P2P trading, while increasing electricity sales revenue by 6.4\% and 44.7\%, respectively. Additionally, peak hour grid demand is reduced by 38.8\% without P2P and 45.6\% with P2P. These improvements are even more pronounced when P2P trading is enabled, highlighting the synergy between advanced forecasting and market mechanisms for resilient, economically efficient energy communities.
\end{abstract}

\end{frontmatter}


\section{Introduction}

\blfootnote{Proceedings of the Main Track of the European Conference on Artificial Intelligence (ECAI 2025), October 25-30, 2025. \url{https://ecai2025.org/}}
The global energy landscape is undergoing a profound transformation driven by the proliferation of distributed energy resources (DERs), the imperative to decarbonize, and the emergence of digital platforms that enable decentralized market participation. Peer-to-peer (P2P) energy trading has rapidly evolved as a promising paradigm, empowering prosumers to directly exchange electricity, optimize local renewable utilization, and contribute to the reduction of carbon emissions in power systems \cite{song2025robust}.

Recent developments in P2P trading frameworks have focused on integrating renewable energy sources, coupling electricity and carbon markets, and leveraging advanced digital infrastructure such as blockchain to ensure transparency and trust \cite{boumaiza2024blockchain} \cite{song2025robust}. These innovations facilitate not only the economic optimization of local energy exchanges but also the explicit accounting and trading of carbon emission allowances, which is increasingly recognized as essential for achieving global climate targets \cite{cheng2018modeling}.

While reinforcement learning (RL) has long addressed sequential decision-making, it faces significant challenges in high-dimensional environments due to the curse of dimensionality, sample inefficiency, and difficulties with sparse rewards and function approximation, resulting in slow convergence and high computational demands \cite{dulac2021challenges}. Advances in deep learning and multi-agent systems have improved RL’s ability to learn optimal policies in complex settings \cite{nguyen2020deep}. Consequently, multi-agent RL (MARL) are increasingly adopted for P2P energy trading in prosumer communities \cite{shah2024multi}.

Complex and dispersed, modern real-world systems have many parts, nonlinear processes, and uncertain environments (\cite{dorri2018multi}). A central challenge in the operation of P2P energy markets is the inherent uncertainty associated with renewable generation and dynamic load profiles. The variability of solar and wind resources, as well as the stochastic nature of consumer demand, introduce significant risks that can undermine both economic efficiency and system reliability if not properly managed \cite{guo2020chance} \cite{hu2020bayesian}. Traditional deterministic forecasting approaches are insufficient in this context, as they fail to capture the full spectrum of possible future scenarios, leading to suboptimal or risk-prone trading and dispatch decisions.

To address these challenges, recent research has emphasized the importance of robust and uncertainty-aware forecasting methods. Probabilistic forecasting, which provides not only point estimates but also confidence intervals or probability distributions, enables market participants to make risk-informed decisions and supports the design of resilient trading mechanisms \cite{guo2020chance}. Furthermore, the integration of uncertainty quantification into multi-agent optimization and reinforcement learning frameworks has been shown to enhance the adaptability and robustness of P2P trading systems, particularly in the presence of high renewable penetration and carbon constraints \cite{song2025robust}.

In parallel, the coupling of energy and carbon markets within P2P trading platforms is gaining traction as a means to internalize the environmental externalities of electricity consumption and incentivize low-carbon behaviors \cite{cheng2018modeling}. By enabling the joint trading of electricity and carbon emission allowances, these systems can more effectively align individual prosumer incentives with broader decarbonization objectives.

This paper addresses these emerging needs by proposing a novel framework that integrates a heteroscedastic probabilistic transformer-based prediction model, Knowledge Transformer with Uncertainty (KTU), with MARL for advanced P2P energy and carbon trading. The approach explicitly models uncertainty in both load and renewable generation, propagates this information into trading and dispatch decisions, and evaluates the resulting impacts on economic performance and carbon emissions. Through this integration, the state of the art in resilient, efficient, and sustainable P2P energy systems is advanced.

\section{Related Work}

P2P energy trading in microgrids has emerged as a decentralized approach to sustainable energy distribution, facing challenges in scalability, privacy, pricing, and uncertainty. Zhou et al. demonstrated that early community market mechanisms apply uniform prices, limiting individualized incentives~\cite{zhou2018evaluation}. While Zheng et al. introduced auction-based methods for trader-specific pricing~\cite{zheng2024multi}, these approaches struggle with real-world uncertainties in trader behavior and energy supply.

May et al. presented MARL as a promising solution, showing how agents can learn optimal strategies in dynamic environments~\cite{may2023multi}. Bhavana et al. identified persistent technical challenges regarding scalability and uncertainty management~\cite{bhavana2024applications}, while Bassey et al. investigated AI applications in trading strategy optimization~\cite{bassey2024peer}. However, most implementations rely on deterministic forecasts, inadequately capturing the inherent variability in renewable systems. Zhang et al. demonstrated that forecasting errors significantly impact market efficiency~\cite{zhang2024assessment}, highlighting the need for uncertainty-aware forecasting models.

In transformer architectures, Liu et al. have shown promising results in energy forecasting~\cite{liu2023itransformer}, but the approach primarily addresses single-agent settings or deterministic outputs. Chen et al. developed a DQN-based approach for price prediction~\cite{chen2019realistic}, though without uncertainty quantification. El et al. investigated uncertainty-aware prosumer coalitional games~\cite{el2017managing}, but did not integrate probabilistic forecasting with multi-agent learning.

Recent work by Yazdani et al. proposed robust optimization for real-time trading~\cite{yazdani2023forecast}, while Uthayansuthi et al. combined clustering, forecasting, and deep reinforcement learning~\cite{uthayansuthi2024optimization}. However, these approaches either lack advanced neural forecasting integration or focus primarily on economic optimization without considering uncertainty impact.

This work advances the state-of-the-art through several key innovations. The Knowledge Transformer with Uncertainty model is introduced for probabilistic forecasts with heteroscedastic uncertainty, specifically tailored for P2P energy trading. Unlike previous approaches, the proposed framework integrates uncertainty-aware forecasts directly into a multi-agent DQN-based MARL system, enabling risk-sensitive trading decisions. Additionally, carbon accounting as peak tariff management is incorporated into the reward structure, addressing both economic and environmental objectives. The integration of automated hyperparameter optimization further distinguishes this approach, ensuring optimal performance across both forecasting and trading modules.

\section{Models \& Architecture}

\subsection{Data and Feature Engineering}

The P2P energy trading community consists of 10 rural Finnish prosumers, each equipped with PV and battery systems: 4 dairy farms (data from Uski et al.~\cite{uski2018microgrid}) and 6 households (synthetic loads based on Finnish profiles and seasonal multipliers~\cite{fingrid_load_generation,statfi_energy}), with 2 households owning EVs. PV generation is simulated using SAM~\cite{samNrel}. In line with~\cite{SEAI2024}, renewable capacities are set to 40\% of the annual load.

Robust feature engineering underpins accurate P2P energy forecasting~\cite{von2018distributed}. Multi-prosumer data are aggregated and normalized, prosumer size is encoded categorically, and cyclical (sine/cosine) as well as one-hot encodings are applied for temporal and seasonal features~\cite{lim2021temporal}. To capture high-latitude solar patterns, a custom daylight feature is derived from Helsinki’s astronomical data. Supervised learning sequences are constructed via a sliding window, following transformer-based time-series best practices~\cite{wu2021autoformer}.

\begin{figure}[t]
    \centering
    \includegraphics[width=\columnwidth]{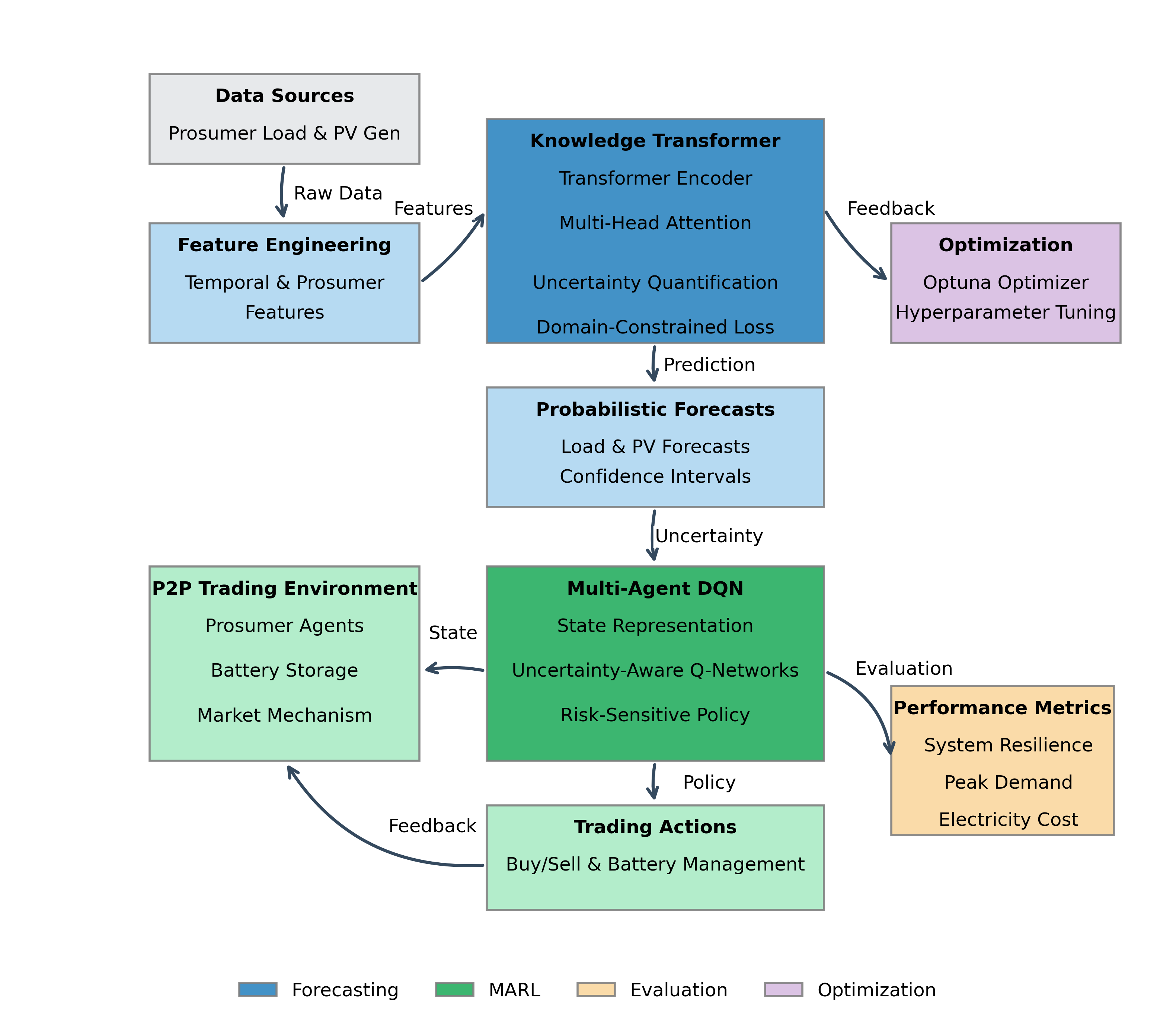}
    \caption{KTU-DQN Ensemble Architecture for P2P Energy Trading}
    \label{fig:model architecture}
\end{figure}

\subsection{Knowledge Transformer with Uncertainty (KTU) Model}

The KTU (Knowledge-driven Transformer with Uncertainty) model is a heteroscedastic probabilistic transformer for energy forecasting, building upon recent developments in transformer-based architectures~\cite{salinas2020deepar,vaswani2017attention}. An overview of the model architecture is illustrated in Figure~\ref{fig:model architecture}. The architecture comprises an input projection layer with learnable positional encodings~\cite{zhou2021informer}, followed by a transformer encoder that utilizes multi-head self-attention to model temporal dependencies~\cite{zhou2022fedformer}. Specifically, the multi-head attention mechanism is defined as
\begin{equation}
    \mathbf{H} = \text{MultiHead}(\mathbf{Q}, \mathbf{K}, \mathbf{V}) = \text{Concat}(head_1,\ldots,head_h)\mathbf{W}^O
\end{equation}
where $\mathbf{Q}$, $\mathbf{K}$, and $\mathbf{V}$ denote the query, key, and value matrices, respectively, and $\mathbf{W}^O$ is the output projection matrix. Each attention head $head_i$ is computed as
\begin{equation}
    head_i = \text{Attention}(\mathbf{Q}\mathbf{W}_i^Q, \mathbf{K}\mathbf{W}_i^K, \mathbf{V}\mathbf{W}_i^V)
\end{equation}
where $\mathbf{W}_i^Q$, $\mathbf{W}_i^K$, and $\mathbf{W}_i^V$ are learnable projection matrices for the $i$-th head. This enables the model to capture complex temporal patterns across multiple representation subspaces.

The KTU model features dual output heads that predict both the mean $\mu$ and variance $\sigma^2$ (via Softplus activation) for each target, thereby capturing aleatoric uncertainty~\cite{pearce2020uncertainty}. The model outputs probabilistic predictions for both prosumer load $L$ and photovoltaic (PV) generation $P$ as follows:
\begin{equation}
    \begin{split}
        p(L_{t+k}|\mathbf{x}_t) &= \mathcal{N}(\mu_L(\mathbf{x}_t), \sigma^2_L(\mathbf{x}_t)) \\
        p(P_{t+k}|\mathbf{x}_t) &= \mathcal{N}(\mu_P(\mathbf{x}_t), \sigma^2_P(\mathbf{x}_t))
    \end{split}
\end{equation}
where $\mathbf{x}_t$ denotes the input features at time $t$, $L_{t+k}$ and $P_{t+k}$ are the load and PV generation at forecast horizon $t+k$, and $\mu_L(\mathbf{x}_t)$, $\sigma^2_L(\mathbf{x}_t)$, $\mu_P(\mathbf{x}_t)$, $\sigma^2_P(\mathbf{x}_t)$ are the predicted means and variances for load and PV, respectively.

To ensure physical plausibility, the PV mean prediction is modulated by physics-informed constraints based on daylight and seasonality:
\begin{equation}
    \mu_P^{\text{final}}(\mathbf{x}_t) = \text{softplus}(\mu_P(\mathbf{x}_t)) \cdot x_t^{\text{daylight}} \cdot x_t^{\text{norm\_daylight}}
\end{equation}
where $x_t^{\text{daylight}}$ is a binary indicator of daylight presence (1 if daytime, 0 if night), and $x_t^{\text{norm\_daylight}}$ encodes the normalized daylight duration for the given season~\cite{karniadakis2021physics}.

The model is optimized using Optuna~\cite{akiba2019optuna} for hyperparameter tuning, targeting a three-hour-ahead joint forecast of prosumer load and PV generation. The architecture implements a two-layer feedforward network projecting inputs to a 128-dimensional space with layer normalization and ReLU activation~\cite{wu2021autoformer}. Two transformer encoder layers, each comprising four attention heads and 512-dimensional feedforward networks with 0.1 dropout, process the encoded inputs. Dual output heads with Softplus activation for variance prediction generate the final forecasts. Key hyperparameters include learning rate, batch size, hidden dimensions, number of attention heads $h$, dropout rate, and regularization weights.

The model employs a composite loss function that combines the Gaussian negative log-likelihood with domain-specific regularization terms:
\begin{equation}
    \begin{split}
        \mathcal{L} = &\frac{1}{2}\sum_{i=1}^N \left[\log(\sigma^2_i + \epsilon) + \frac{(y_i - \mu_i)^2}{\sigma^2_i + \epsilon}\right] \\
        &+ \alpha\sum_{t=1}^{T-1}|\mu_{t+1} - \mu_t| \\
        &+ \beta\sum_{t=1}^T P_t(1 - x_t^{\text{daylight}})
    \end{split}
\end{equation}
where $N$ is the number of training samples, $y_i$ is the ground truth target, $\mu_i$ and $\sigma^2_i$ are the predicted mean and variance for sample $i$, $\epsilon$ is a small constant for numerical stability, $\alpha$ controls the strength of the temporal smoothness regularization, $\beta$ penalizes physically impossible nighttime PV generation, $T$ is the sequence length, and $P_t$ is the predicted PV generation at time $t$.

For probabilistic forecasting, samples are drawn from the predicted mean and variance to construct empirical confidence intervals. Evaluation metrics include Prediction Interval Coverage Probability (PICP), Mean Prediction Interval Width (MPIW), and Continuous Ranked Probability Score (CRPS), which assess the quality and calibration of the probabilistic forecasts.

\subsection{Deep Q-Networks}

Traditional Q-learning employs a Q-table to learn optimal policies, but this approach is impractical for large or continuous state-action spaces due to memory constraints and poor generalization. Deep Q-Networks (DQN) address this by using deep neural networks to approximate Q-values, enabling effective learning in high-dimensional environments~\cite{lv2019path}. 

DQN extends Q-learning by replacing the Q-table with a neural network parameterized by $\theta$, which is updated via stochastic gradient descent to minimize the temporal-difference (TD) error:
\begin{align}
\theta_{t+1} =\ & \theta_t + \alpha \Big[ r_t + \gamma \max_{a'} q(s_{t+1}, a'; \theta_{\text{TD}}) \notag \\
& \qquad -\ q(s_t, a_t; \theta_t) \Big] \nabla_\theta q(s_t, a_t; \theta_t)
\end{align}
Here, $\theta_{\text{TD}}$ denotes the target network, periodically updated to stabilize training. This framework allows DQN to efficiently learn policies in complex, high-dimensional spaces where traditional Q-learning fails.

\subsection{Pricing Model and Double Auction}

The system employs a distributed P2P energy trading model, where only energy generation and load data are shared with a centralized auctioneer. Market clearance is performed centrally, but each participant independently manages their load, generation, and battery, preserving privacy by withholding additional information from external entities.

Participants report their electricity surplus or deficit to a centralized agent, which serves as both auctioneer and advisor. This agent evaluates market conditions and determines the Internal Selling Price (ISP) and Internal Buying Price (IBP) using the Supply and Demand Ratio (SDR) method, enabling real-time price setting based on current system demand and supply~\cite{liu2017energy}. The ISP is the price for selling surplus energy within the community, while the IBP is the price for purchasing energy, ensuring fair and transparent transactions.

This study employs a double auction (DA) mechanism, adapted from Qiu et al.~\cite{qiu2022mean}. The auctioneer requires only load, generation, and pricing data from participants, thereby preserving data privacy. The auction system comprises buyers, $\beta$, and sellers, $\sigma$. Each buyer $b$ submits a bid with price $P_{\beta,b}$ and quantity $Q_{\beta,b}$, while each seller $s$ submits an offer with price $P_{\sigma,s}$ and quantity $Q_{\sigma,s}$. The auctioneer maintains two order books: $O_b$ for buy orders and $O_s$ for sell orders, both sorted by price. Based on the ISP and IBP, each agent submits bids or offers specifying the price and quantity of electricity to trade. The auctioneer clears the market using the auction algorithm, ensuring that energy is allocated efficiently and that transactions remain fair and privacy-preserving for all participants.

\section{Proposed Approach}

\begin{figure*}[t]
    \centering
    \includegraphics[width=1\textwidth]{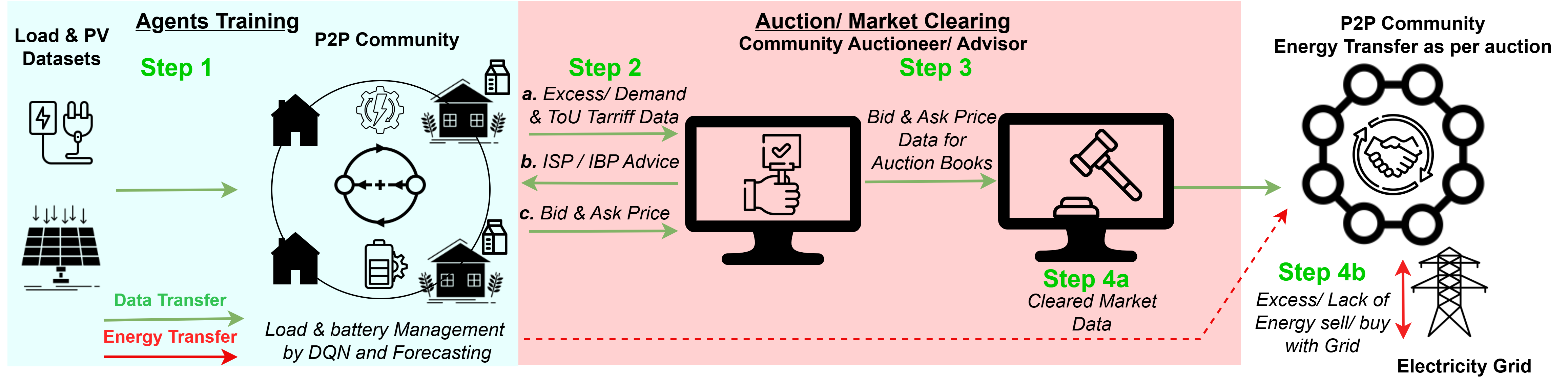}
    \caption{Process flow of Uncertainty-aware Forecasting DQN simulator}
    \label{fig:Methodology}
\end{figure*}

The uncertainty-aware forecasting enabled MARL P2P energy trading model simulates 10 prosumer agents over 2 million timesteps using the PettingZoo framework. Although P2P participants are self-interested, MARL remains realistic as agents are trained independently to maximize their own utility, without explicit coordination. This allows each agent to adapt to the dynamic actions of others, effectively modeling decentralized, competitive P2P energy trading environments. Each agent’s state space includes current and forecasted load, generation, battery status, and uncertainty estimates from the probabilistic forecasting model. The action space consists of discrete actions representing energy management strategies such as buying, selling, charging, discharging, and self-consumption. The reward function is designed for each action to incorporate forecast uncertainty, tariff periods, and battery constraints (see equations in the Reward Function section).

\begin{algorithm}
\caption{Uncertainty-Aware MARL for P2P Energy Trading}
\label{pseudocode}
\begin{algorithmic}[1]
\State Initialize: agent set $N$, battery capacities $B_{cap}$, time $t=0$
\While{$t \leq T$ (simulation period)}
    \If{end of day} 
        \State Reset hour counter and increment day
    \EndIf
    \For{each agent $i \in N$}
        \State Observe current $(L_{i,t}, G_{i,t}, B_{i,t})$
        \State Get forecasts $(FL_{i,t}, FG_{i,t})$ and uncertainties $(U_{L,i,t}, U_{G,i,t})$
        \State Form state vector $s_{i,t} = [L_{i,t}, G_{i,t}, B_{i,t}, FL_{i,t}, FG_{i,t}, U_{L,i,t}, U_{G,i,t}]$
        \State Select action $a_{i,t}$ using DQN policy $\pi(s_{i,t})$
        \State Calculate energy balance $E_{i,t} = G_{i,t} - L_{i,t}$
        \If{$E_{i,t} < 0$ or $a_{i,t}$ is buy}
            \State Add to BuyerBook$(i, |E_{i,t}|, p_{bid})$
        \ElsIf{$E_{i,t} > 0$ or $a_{i,t}$ is sell}
            \State Add to SellerBook$(i, E_{i,t}, p_{ask})$
        \EndIf
    \EndFor
    \State \textbf{Market Clearing:}
    \State Calculate SDR = $\sum Supply / \sum Demand$
    \If{$0 \leq SDR \leq 1$}
        \State Calculate ISP = $\frac{\lambda_{sell} \lambda_{buy}}{(\lambda_{buy} - \lambda_{sell})SDR + \lambda_{sell}}$
        \State Calculate IBP = $ISP \cdot SDR + \lambda_{buy} \cdot (1-SDR)$
    \EndIf
    \State Match buyers and sellers by price priority using ISP and IBP
    \State Update rewards $R_{i,t}$ and advance time $t \gets t + 1$
\EndWhile
\end{algorithmic}
\end{algorithm}

Agents are trained individually using DQN, and their models are evaluated over 10 independent episodes, with results averaged to ensure robustness. After training, agent actions are aggregated to compute hourly buy/sell quantities and battery levels. A DA mechanism clears the market at each timestep: agents submit bids and asks based on their needs and internal price signals, and trades are matched to maximize local exchange before resorting to grid transactions. Dynamic pricing is determined by SDR and grid tariffs, ensuring realistic market behavior. The overall workflow and system architecture are illustrated in Figure~\ref{fig:Methodology}. A step-by-step description of the system is provided in Algorithm~\ref{pseudocode}. The P2P network consists of $N$ agents, each with battery capacity $B_{cap}$. At time $t$, agent $i$ observes its current load ($L_{i,t}$), generation ($G_{i,t}$), and battery state ($B_{i,t}$). The knowledge transformer provides load and generation forecasts ($FL_{i,t}$, $FG_{i,t}$) with associated uncertainties ($U_{L,i,t}$, $U_{G,i,t}$). These values form the state vector $s_{i,t}$, which is used by the DQN policy $\pi$ to select an action $a_{i,t}$. The energy balance $E_{i,t}$ determines whether the agent acts as a buyer (with bid price $p_{bid}$) or a seller (with ask price $p_{ask}$). During market clearing, the SDR is calculated and used to determine the ISP and IBP. These prices are used by the prosumers to ask or bid accordingly. At the end, buyers and sellers are matched, trades are executed, and agent rewards are updated $R_{i,t}$ before advancing to the next timestep.

\subsection*{Reward Functions with Uncertainty-aware Forecasting}

\textbf{Notation:}
\begin{itemize}
    \item $G_t^i, L_t^i$: Generation and load
    \item $\text{SoC}_t^i$: State of charge (\%)
    \item $T_{\text{grid}}$: Grid tariff $\in \{N, NP, P, D\}$
    \item $\hat{G}_{t+k}^i, \hat{L}_{t+k}^i$: Future forecasts
    \item $\Delta_{\text{peak}}^i$: Peak deficit prediction
    \item $\alpha_t^i$: Confidence score
\end{itemize}

\begin{enumerate}
    \item \textbf{Charge and Buy}
    \[
    R = 
    \begin{cases}
    0.5 + 1.5\alpha_t^i + 1.0, & \text{if } \phi_1 \\
    0.5 + \alpha_t^i, & \text{if } \phi_2 \\
    0.5, & \text{if } \phi_3 \\
    0, & \text{if } T_{\text{grid}} = P
    \end{cases}
    \]
    where:\\
    $\phi_1: \text{SoC}_t^i \leq 90\% \land T_{\text{grid}} = NP \land \Delta_{\text{peak}}^i > 0$\\
    $\phi_2: \text{SoC}_t^i \leq 90\% \land T_{\text{grid}} = N$\\
    $\phi_3: \text{SoC}_t^i \leq 90\% \land G_t^i < L_t^i$

    \item \textbf{Buy}
    \[
    R = 
    \begin{cases}
    0.25, & \text{if } \phi_1 \land T_{\text{grid}} = P \\
    0.5, & \text{if } \phi_1 \\
    0, & \text{otherwise}
    \end{cases}
    \]
    where $\phi_1: G_t^i < L_t^i \land \text{SoC}_t^i < 10\%$

    \item \textbf{Sell}
    \[
    R = 
    \begin{cases}
    0.75, & \text{if } \phi_1 \land T_{\text{grid}} = P \\
    0.5, & \text{if } \phi_1 \\
    0, & \text{otherwise}
    \end{cases}
    \]
    where $\phi_1: G_t^i > L_t^i \land (\text{SoC}_t^i \geq 90\%)$

    \item \textbf{Discharge and Sell}
    \[
    R = 
    \begin{cases}
    (0.5 + 0.5\alpha_t^i)1.5, & \text{if } \phi_1 \\
    0.5, & \text{if } \phi_2 \\
    0, & \text{otherwise}
    \end{cases}
    \]
    where:\\
    $\phi_1: G_t^i > L_t^i \land \text{SoC}_t^i \geq 20\% \land T_{\text{grid}} = P$\\
    $\phi_2: G_t^i > L_t^i \land \text{SoC}_t^i \geq 90\%$

    \item \textbf{Discharge and Buy}
    \[
    R = 
    \begin{cases}
    (0.5 + 0.5\alpha_t^i)1.5, & \text{if } \phi_1 \land T_{\text{grid}} = P \\
    0.5, & \text{if } \phi_1 \\
    0, & \text{otherwise}
    \end{cases}
    \]
    where $\phi_1: G_t^i < L_t^i \land \text{SoC}_t^i \geq 10\%$

    \item \textbf{Self-Consumption}
    \[
    R = 
    \begin{cases}
    1.2, & \text{if } \phi_1 \land T_{\text{grid}} = P \\
    1.0, & \text{if } \phi_1 \\
    0.5, & \text{if } \phi_2 \\
    0, & \text{otherwise}
    \end{cases}
    \]
    where:\\
    $\phi_1: |G_t^i - L_t^i| \leq 0.1$\\
    $\phi_2: 0.1 < |G_t^i - L_t^i| \leq 0.2$

    \item \textbf{Self and Charge}
    \[
    R = 
    \begin{cases}
    0.5 + 2.0\alpha_t^i + 1.0, & \text{if } \phi_1 \\
    0.5 + 0.5\alpha_t^i, & \text{if } \phi_2 \\
    0, & \text{if } T_{\text{grid}} = P
    \end{cases}
    \]
    where:\\
    $\phi_1: G_t^i > L_t^i \land \text{SoC}_t^i \leq 90\% \land T_{\text{grid}} = NP$\\
    $\phi_2: G_t^i > L_t^i \land \text{SoC}_t^i \leq 90\%$

    \item \textbf{Self and Discharge}
    \[
    R = 
    \begin{cases}
    (0.5 + 0.5\alpha_t^i)1.5, & \text{if } \phi_1 \land T_{\text{grid}} = P \\
    0.5, & \text{if } \phi_1 \\
    0, & \text{otherwise}
    \end{cases}
    \]
    where $\phi_1: G_t^i < L_t^i \land \text{SoC}_t^i \geq 20\%$
\end{enumerate}

\begin{figure}[t]
    \centering
    \includegraphics[width=\columnwidth]{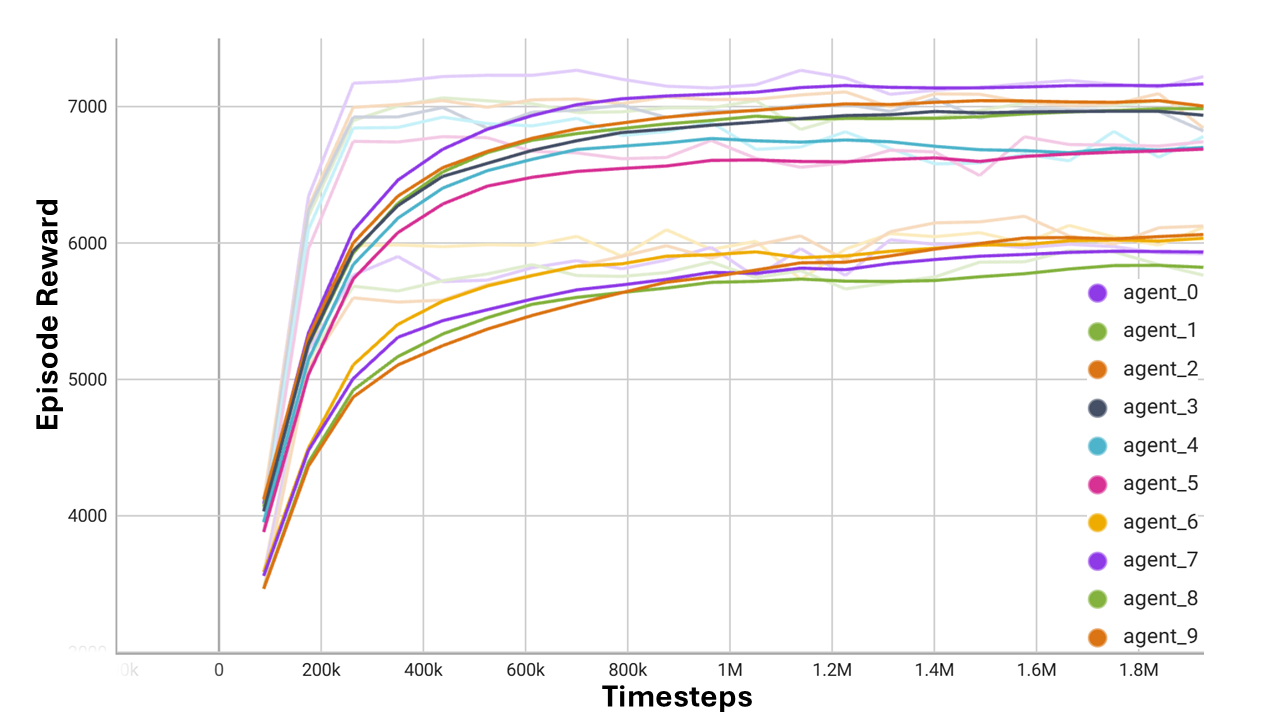}
    \caption{Reward Convergence over 2M time steps}
    \label{fig:convergence}
\end{figure}

\begin{table*}[ht]
\caption{Evaluation of KPIs for Various Models With and Without P2P Trading}
\centering
\begin{tabular*}{\textwidth}{@{\extracolsep{\fill}}lcccccc@{}}
\hline
\textbf{Metric} & \textbf{Scenario} & \textbf{Rule Based} & \textbf{RB+QL} & \textbf{DQN} & \textbf{DQN Forecasting} & \textbf{\% Diff (DQN vs DQN Forecasting)} \\
\hline
\multirow{3}{*}{Electricity Cost (Bought) (€)} 
  & w/o P2P & 125400 & 121300 & 105000 & \textbf{99100} & \textbf{-5.7\%} \\
  & with P2P & 119500 & 116800 & 102100 & \textbf{96800} & \textbf{-3.2\%} \\
  & P2P vs w/o P2P (\%) & -4.7\% & -3.7\% & -2.8\% & \textbf{-2.9\%} & \\
\hline
\multirow{3}{*}{Electricity Revenue (Sold) (€)} 
  & w/o P2P & 3600 & 3800 & 7850 & \textbf{8350} & \textbf{+6.4\%} \\
  & with P2P & 4400 & 4650 & 14450 & \textbf{20900} & \textbf{+44.7\%} \\
  & P2P vs w/o P2P (\%) & +22.2\% & +22.4\% & +84.1\% & \textbf{+150.1\%} & \\
\hline
\multirow{3}{*}{Peak Hour Demand (kW)} 
  & w/o P2P & 36000 & 34500 & 23200 & \textbf{14200} & \textbf{-38.8\%} \\
  & with P2P & 28500 & 26600 & 21850 & \textbf{11900} & \textbf{-45.6\%} \\
  & P2P vs w/o P2P (\%) & -20.8\% & -22.9\% & -5.8\% & \textbf{-16.2\%} & \\
\hline
\end{tabular*}
\label{results_comparison}
\end{table*}

\section{Evaluation \& Discussion}

The integration of uncertainty-aware forecasting into the MARL DQN framework yields substantial improvements over traditional approaches in P2P energy trading. Figure~\ref{fig:convergence} shows the episode reward trajectories for 10 agents over 1.8 million time steps.  All agents experience a rapid increase in episode reward during early training and converges at around 600k steps. Notably, the proposed uncertainty-aware forecasting DQN achieves convergence approximately 50\% faster than the standard DQN, requiring about 25\% fewer time steps to reach high performance. This efficiency is attributed to the model’s ability to anticipate future states using probabilistic forecasts, which narrows the exploration space and guides agents toward optimal policies more effectively. In practical microgrids, participant numbers commonly range from a few dozen to several hundred \cite{griego2019aggregation}. The modular and parallelizable environment, featuring a linear-complexity auction and communication-free agents, enables straightforward scaling for large, distributed systems. This approach combines decentralized P2P load and battery management with a centralized auction, prioritizing scalability and resilience over the optimality of negotiation-based MARL, while leveraging uncertainty-aware forecasts for informed decisions. MARL training can further support larger populations through distributed or federated learning.

\begin{figure}[t]
    \centering
    \includegraphics[width=\columnwidth]{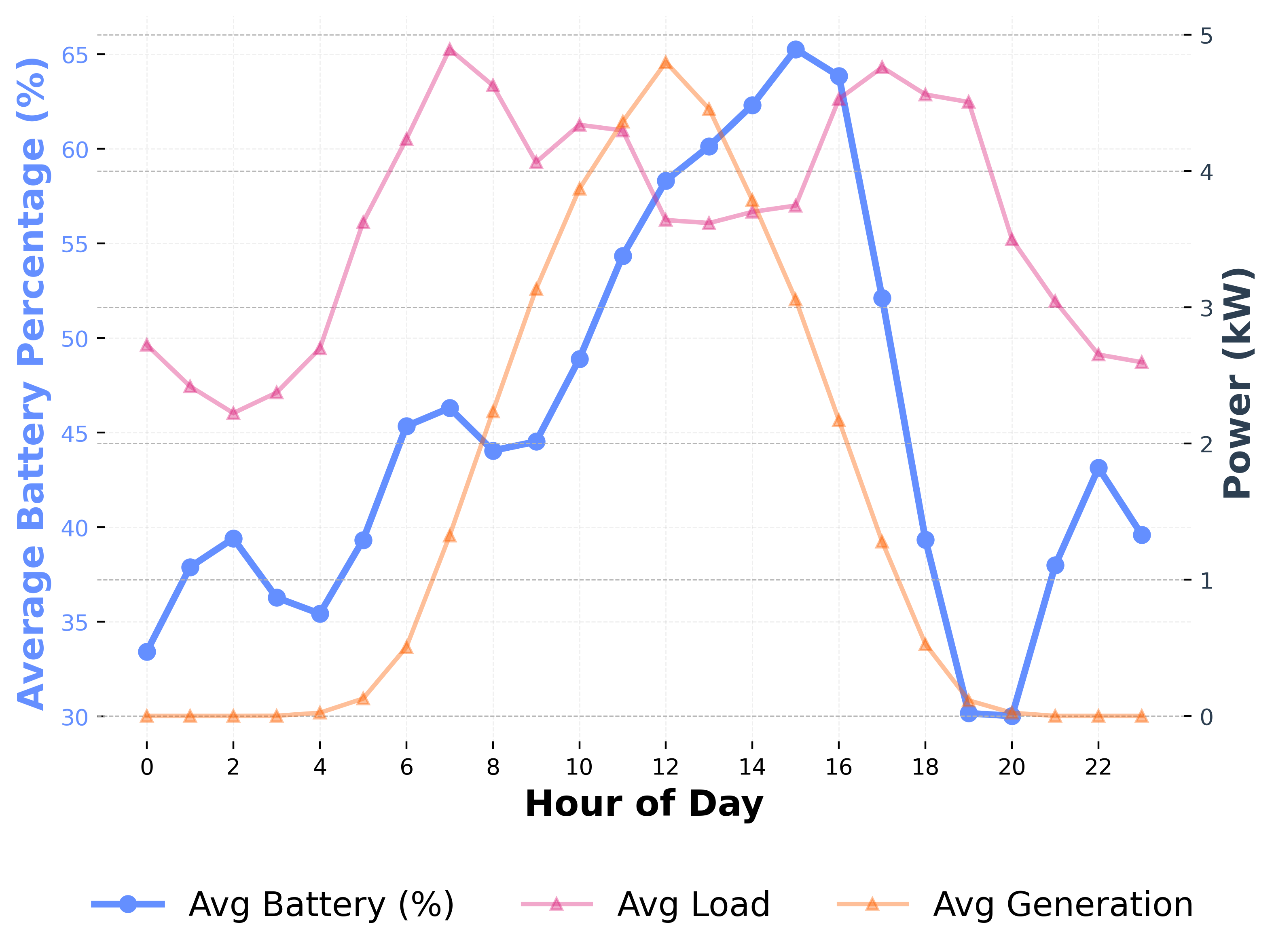}
    \caption{Daily Battery SOC, Load \& Generation (Year Avg.)}
    \label{fig:dqn_battery}
\end{figure}

Battery management is also significantly improved, as illustrated in Figure~\ref{fig:dqn_battery}. The average battery percentage rises steadily from early morning, peaking in the late afternoon, and then declines as stored energy is dispatched to meet evening loads. The battery is usually pre-charged before the peak hours start in the evening, which is incorporated to reduce the reliance on the utility grid during peak tariff hours for cost effectiveness and to lower carbon emissions associated with peak hour generation. This pattern demonstrates that agents, informed by uncertainty-aware forecasts, coordinate charging during periods of high renewable generation and discharging during peak demand. Such anticipatory behavior contrasts with the more reactive strategies of standard DQN and rule-based methods, leading to more effective and community-beneficial storage utilization.

Key performance indicators (KPIs) further highlight the advantages of the proposed approach in Table \ref{results_comparison}. In terms of \textbf{energy purchase cost}, the uncertainty-aware DQN reduces costs by approximately \textbf{5.7\%} without P2P trading (from~€105{,}000 to~€99{,}100) and by \textbf{3.2\%} with P2P trading (from~€102{,}100 to~€96{,}800) compared to the standard DQN. For \textbf{revenue from electricity sales}, the proposed model achieves a \textbf{6.4\%} increase without P2P (from~€7{,}850 to~€8{,}350) and a remarkable \textbf{44.7\%} increase with P2P (from~€14{,}450 to~€20{,}900). Regarding \textbf{peak hour energy demand from the grid}, the uncertainty-aware DQN reduces demand by \textbf{38.8\%} without P2P (from~23{,}200~kW to~14{,}200~kW) and by \textbf{45.6\%} with P2P (from~21{,}850~kW to~11{,}900~kW). These improvements are even more pronounced when P2P trading is enabled, highlighting the synergy between advanced forecasting and market mechanisms.

Although rule-based and ensemble (RB+QL) methods—whose rules and algorithms are described in \cite{shah2024peer, shah2024reinforcement}—provide incremental improvements, they do not offer the adaptability or predictive capabilities of DQN-based approaches. Other MARL algorithms such as PPO were also evaluated, but DQN consistently outperformed them in capturing the complexities of the trading environment and delivering superior economic outcomes. The most significant factor in lowering electricity costs, reducing peak hour demand, and increasing revenue is the implementation of P2P energy trading, which amplifies the benefits of RL models. Overall, the integration of uncertainty-aware forecasting with MARL DQN not only accelerates convergence and improves decision quality but also leads to more effective battery management and superior economic performance, establishing a new benchmark for advanced P2P energy trading systems.


\section{Conclusion}

This paper presented a novel framework that combines uncertainty-aware knowledge transformer forecasting with multi-agent deep reinforcement learning for advanced P2P energy trading. By enabling agents to make risk-sensitive, anticipatory decisions, the proposed approach achieves up to 50\% faster convergence and more efficient battery management, as agents coordinate charging and discharging based on forecasted generation and load.

Key performance indicators underscore the practical benefits: the uncertainty-aware DQN reduces energy purchase costs by 5.7\%, increases revenue from electricity sales by 44.7\% with P2P trading, and lowers peak hour grid demand by 45.6\% compared to standard DQN. These improvements result from the synergy between advanced forecasting and P2P market mechanisms.

Overall, these results set a new benchmark for resilient and efficient P2P energy trading, demonstrating that uncertainty-aware learning is essential for both economic and operational gains in decentralized energy systems. Future work will explore the integration of additional market mechanisms, real-world pilot deployments, optimization of the forecasting horizon, and theoretical analysis or guarantees regarding the convergence properties for even greater impact.


\begin{ack}
This publication is based on research funded by Research Ireland under Grant Number [21/FFP-A/9040].
\end{ack}



\bibliography{mybibfile}

\end{document}